\RequirePackage{fix-cm}
\documentclass[twocolumn]{svjour3}          
\smartqed  
\usepackage{graphicx}
\DeclareMathAlphabet{\mathcal}{OMS}{cmsy}{m}{n}
\usepackage{newtxtext,newtxmath}

\usepackage{algorithmic}
\usepackage{algorithm}
\usepackage{newfloat}

\usepackage{epstopdf}
\usepackage{url}
\usepackage{natbib}
\usepackage{graphicx}
\usepackage{multirow}
\usepackage{color,array}
\usepackage{listings}
\usepackage{xcolor}
\usepackage{colortbl}
\usepackage{amsmath}
\usepackage{pdfpages}
\usepackage{hyperref}
 \usepackage{caption}
\usepackage{makecell}
\usepackage{bm}
\usepackage{tabularx}
\usepackage{hhline}
\usepackage{booktabs}
\usepackage{mathrsfs}
\usepackage{ragged2e}
\usepackage{amsfonts}
\usepackage{times}
\usepackage{helvet}
\usepackage{courier}
\usepackage{tabularx}
\newtheorem{prop}{Proposition}%

\DeclareMathOperator*{\argmin}{\arg\min}

\newcolumntype{C}{>{\centering\arraybackslash}X}
\hypersetup{
    colorlinks=true,
    linkcolor=blue,
    citecolor=blue,
}

\begin{document}

\title{RHINO: Regularizing the Hash-based Implicit Neural Representation
\thanks{The first two authors contributed equally. The work was supported by National Key Research and Development Project of China (2022YFF0902402) and NSFC (T2221003, 62101242, 62022038, 62025108).}
}


\author{Hao Zhu$^*$ \and
        Fengyi Liu$^*$ \and
        Qi Zhang \and Xun Cao \and Zhan Ma
}

\institute{Hao Zhu, Fengyi Liu, Xun Cao, Zhan Ma \at School of Electronic Science and Engineering, Nanjing University, Nanjing, 210023, China\\\email{caoxun@nju.edu.cn, mazhan@nju.edu.cn}
\and Qi Zhang \at Tencent AI Lab, Shenzhen, 518054, China
}

\date{Received: date / Accepted: date}


\maketitle

\abstract{

The use of Implicit Neural Representation (INR) through a hash-table has demonstrated impressive effectiveness and efficiency in characterizing intricate signals. However, current state-of-the-art methods exhibit insufficient regularization, often yielding unreliable and noisy results during interpolations. We find that this issue stems from broken gradient flow between input coordinates and indexed hash-keys, where the chain rule attempts to model discrete hash-keys, rather than the continuous coordinates. To tackle this concern, we introduce RHINO, in which a continuous analytical function is incorporated to facilitate regularization by connecting the input coordinate and the network additionally without modifying the architecture of current hash-based INRs. This connection ensures a seamless backpropagation of gradients from the network's output back to the input coordinates, thereby enhancing regularization. Our experimental results not only showcase the broadened regularization capability across different hash-based INRs like DINER and Instant NGP, but also across a variety of tasks such as image fitting, representation of signed distance functions, and optimization of 5D static / 6D dynamic neural radiance fields. Notably, RHINO outperforms current state-of-the-art techniques in both quality and speed, affirming its superiority.
}



\section{Introduction}\label{sec1}

Implicit neural representation (INR)~\citep{sitzmann2020implicit}, which characterizes a signal by establishing a continuous mapping function between the coordinate and the attribute using a neural network, is drawing intensive attention. Due to its versatile capability of seamlessly incorporating differentiable physical mechanisms, INR holds significant potential for tacking a range of domain-specific inverse problems, particularly in situations where large scale paired datasets are unavailable, ranging from the cross-model media representation/compression~\citep{gao2022objectfolder,strumpler2022implicit} in image processing to reconstruction/rendering~\citep{mildenhall2020nerf,tewari2022advances,zhu2023pyramid} in vision/graphics, from hologram/tomography imaging~\citep{zhu2022dnf,liu2022recovery} in microscopy to meta-surface design~\citep{chen2020physics} in materials, from partial differential equations solver~\citep{raissi2019physics,karniadakis2021physics} in computational math to fluid simulation~\citep{raissi2020hidden} in hydrodynamics. INR is catalyzing a profound transformation in the landscape of signal processing, heralding a paradigm shift that holds immense promise for diverse applications.

\begin{figure*}[!t]
\centering
\includegraphics[width=\linewidth]{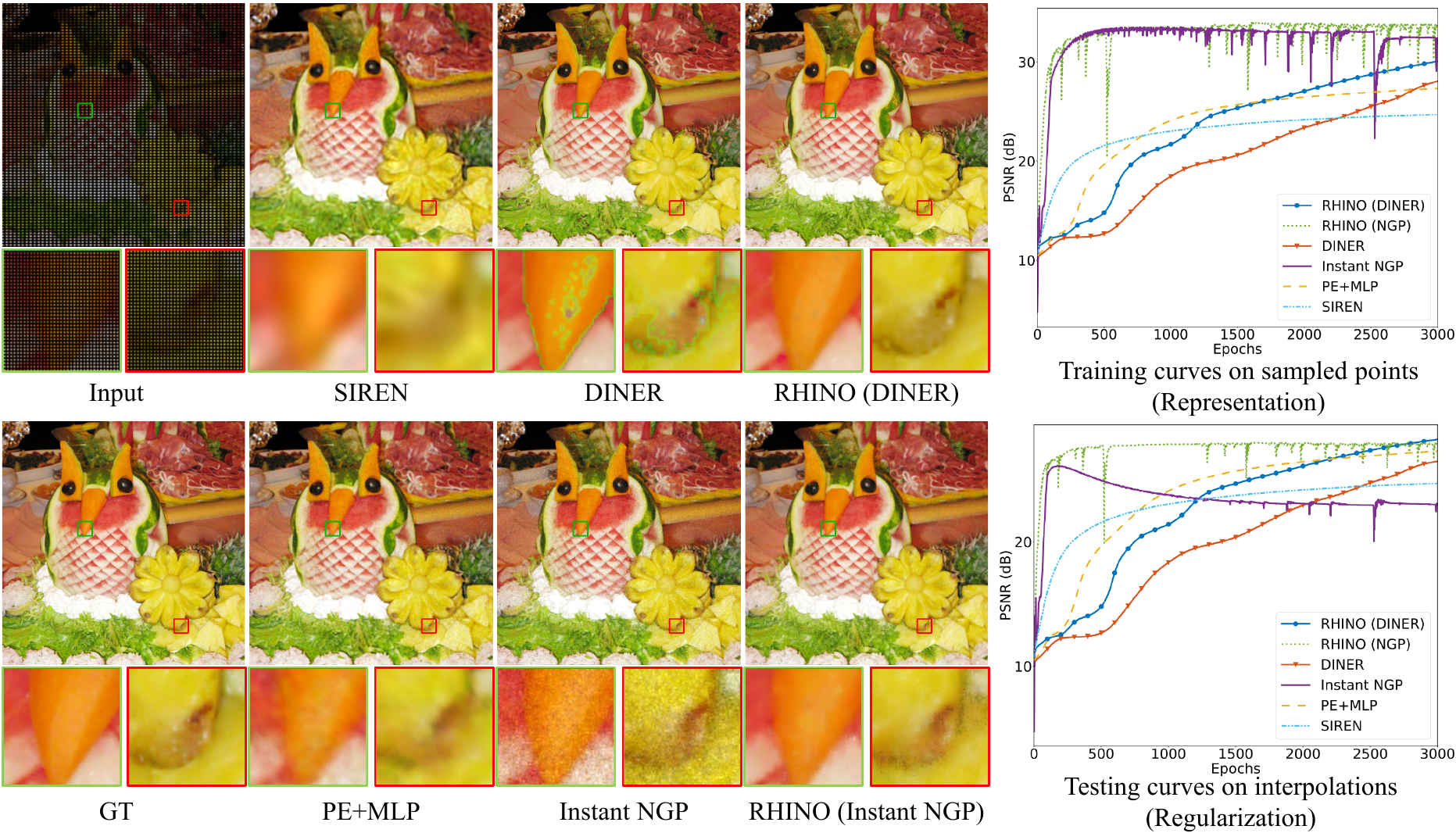} 
\caption{Comparisons of different INRs for representing and interpolating the 2D image 'Pineapple'. Traditional function-expansion-based INRs (SIREN and PE+MLP) suffer from spectral bias, thus over-smooth textures are produced. Hash-based INRs (DINER and Instant NGP) improve the expressive power, however noisy artifacts appear in the interpolations. RHINO improves both the expressive power and the regularization of hash-based INRs.}
\label{fig:res_img_first}
\end{figure*}

However, the expressive power of current INR techniques is limited. For example, they often suffer from the well-known spectral bias~\citep{rahaman2019spectral} that only the information corresponding to the pre-encoded frequencies could be well represented. As a result, it is essential to encode more frequency bases~\citep{tancik2020fourier,sitzmann2020implicit,lindell2022bacon,ramasinghe2022beyond,yuce2022structured,saragadam2023wire} into the neural network at the cost of exceptionally slow convergence. Recently, hash-based methods have garnered significant attention as they circumvent the constraints imposed by spectral bias by replacing the frequency encoding with learnable hash-key indexing~\citep{liu2020neural,takikawa2021neural,chabra2020deep,jiang2020local,muller2022instant,kang2022pixel,xie2023diner}. They have achieved notably higher Peak Signal-to-Noise Ratio (PSNR) values, often even reaching around $\sim$$90$dB~\citep{zhu2023disorder}, for tasks involving signal representation. However, the regularization ability has been considerably weakened or even entirely lost, leading to the emergence of noisy artifacts particularly in tasks that necessitate interpolation.

We have identified a crucial factor responsible for the lost regularization seen in hash-based methods. This phenomenon arises due to a disruption in the gradient flow between the output of the neural network and the input coordinate. The root cause of this issue lies in the indexing of the hash-key, which is based on the relative order of the input coordinate among all coordinates instead of transforming the coordinate using continuous and analytical functions. Consequently, the chain rule, a fundamental principle in calculus, is broken between the input coordinate and the indexed hash-key. This breakage has significant ramifications. The optimization processes for both the hash-keys and the network parameters become detached from the input coordinate. Consequently, the interpolations on coordinates that do not appear in the training process exhibit unwanted artifacts. In light of these challenges, we present an innovative solution named RHINO. This novel approach involves the establishment of an additional continuous connection between the input coordinate and the subsequent network.

RHINO stands as a universal regularization framework designed for a diverse range of hash-based INR backbones. The supplementary connection it introduces holds the potential to substantially bolster regularization performance while concurrently enhancing the system's representation capabilities. Furthermore, the incremental computational overhead incurred remains notably minimal. As shown in Fig.~\ref{fig:res_img_first}, the regularization and representation aspects of two contemporary state-of-the-art methodologies, specifically Instant NGP~\citep{muller2022instant} and DINER~\citep{xie2023diner}, both showcase improvements compared to their original incarnations. Moving forward, to verify the performance of the RHINO on inverse problems, extensive experiments spanning various tasks are conducted, including the 3D signed distance function representation, 5D static, and 6D dynamic neural radiance fields reconstruction. Quantitative and qualitative comparisons with state-of-the-art methods demonstrate the effectiveness and efficiency of the RHINO. Specifically, the main contributions of the work include, 
\begin{enumerate}
    \item We observe and analyse the phenomenon that current hash-based INRs suffer a notable lack of regularization, resulting in noisy artifacts on interpolated points.
    \item We propose a universal regularization framework for hash-based INRs, which serves to enhance both regularization and representation performance while incurring only a minor increase in computational time.
    \item We substantiate that RHINO surpasses prior function-expansion-based and hash-based INRs in the domains of 2D image fitting, and 3D shape representation, as well as 5D static and 6D dynamic neural rendering.
\end{enumerate}

\section{Related work}
This section gives a brief review of existing methods for implicit neural representations. RHINO draws inspiration from hash-based representations and the regularization of INR, we outline some of the related works here to set the context.
\label{sec:relatedwork}
\subsection{Neural Scene Representation and Rendering.}
The use of neural network to represent the geometry and appearance of a scene has witnessed a significant upsurge in popularity.
Traditional methods explicitly represent scenes using point clouds \citep{pumarola2020c, wu2020multi,ruckert2022adop}, meshes \citep{wang2018pixel2mesh,thies2019deferred,huang2020adversarial}, or voxels \citep{wang2017cnn, lombardi2019neural, sitzmann2019deepvoxels}. In contrast, neural network promise 3D-structure-aware and memory-economic scene representations for radiance fields \citep{mildenhall2020nerf, barron2021mip, muller2022instant}, signal distance fields (SDF) \citep{chabra2020deep,yariv2021volume, genova2020local}, or occupancy networks of objects \citep{mescheder2019occupancy, peng2020convolutional, niemeyer2019occupancy}.

A remarkable advancement in this field is neural radiance fields (NeRF), which learns a continuous volumetric representation of a 3D scene from a set of images, enabling the rendering of photo-realistic views through ray tracing. NeRF has recently been investigated for various tasks, such as view synthesis for dynamic scenes \citep{li2022neural, wang2022mixed, fang2022fast, fridovich2023k,li2023dynibar}, implicit surface reconstruction \citep{wang2021neus, yariv2021volume, yu2022monosdf}, generalizable image-based rendering \citep{wang2021ibrnet, chen2021mvsnerf, johari2022geonerf, huang2023local}, and scene editing \citep{martin2021nerf, yuan2022nerf, bao2023sine}. Recently, 2D generative image models (\textit{i.e.}, Generative Adversarial Networks (GANs) \citep{goodfellow2014generative, abdal2019image2stylegan} and diffusion-based methods \citep{ho2020denoising,nichol2021improved,saharia2022photorealistic}) have been extended and combined with scene representations to enable 3D-aware generation, such as EG3D \citep{chan2022efficient}, StyleSDF \citep{or2022stylesdf}, DreamFusion \citep{poole2022dreamfusion}, and Magic3D \citep{lin2023magic3d}. Leveraging the benefits of neural scene representations, these approaches have shown remarkable potential in producing realistic and intricate 3D scenes while preserving novel view consistency. 

\subsection{Implicit Neural Representations}
Implicit neural representations (INRs), the core component for neural scene representations, are designed to learn continuous functions based on a multi-layer perceptron (MLP) that maps coordinates to visual signals, such as images \citep{dupont2021coin, sitzmann2020implicit, tancik2020fourier, lindell2021bacon}, videos \citep{kasten2021layered}, and 3D scenes \citep{martin2021nerf, wang2021neus}. With the widespread application in novel view synthesis \citep{mildenhall2020nerf}, INRs have rapidly expanded into various fields of vision and signal processing, such as cross-model media representation/compression \citep{gao2022objectfolder, strumpler2022implicit}, neural camera representations \citep{huang2022hdr, huang2023inverting}, microscopy imaging \citep{zhu2022dnf, liu2022recovery} and partial differential equations solver~\citep{raissi2019physics, karniadakis2021physics}.

In summary, Existing INRs can be roughly classified into two groups: a) function-expansion-based INRs, and b) hash-based INRs. 
Specifically, \cite{tancik2020fourier} propose a coordinate-based INR that uses Fourier features of coordinates as input of an MLP to learn high-frequency information of natural signals. \cite{sitzmann2020implicit} employ periodic activation function (SIREN) as a replacement for the traditional ReLU activation in INRs, enabling fine details of the scene. Recent works reveal that the success of the Fourier features is achieved by viewing the INR as the problem function-expansion using different bases, as a result, classical bases in higher mathematics such as the wavelet bases~\citep{fathony2020multiplicative,saragadam2023wire}, polynomial bases~\citep{yang2022polynomial} and Gaussian bases~\citep{ramasinghe2022beyond} have all been successfully integrated into the architecture design of INR. Despite the growing interest and success of coordinate-based INRs, existing techniques often face limitations in their expressive power. Issues such as spectral bias and the need for additional frequency bases for complex scenes continue to pose challenges~\citep{yuce2022structured}. 


To overcome the limitations posed by the spectral bias and to amplify the expressive capabilities of INRs, recent techniques \citep{liu2020neural, takikawa2021neural, chabra2020deep, jiang2020local, muller2022instant, yu2022monosdf, kang2022pixel, xie2023diner, sun2022direct, chan2022efficient, chen2022tensorf} have introduced hash-based INRs. These innovative approaches replace frequency encoding with learned hash-key indexing, thereby bestowing these models with significantly enhanced scene representation capabilities. In particular, \cite{muller2022instant} make use of learned hash functions to map coordinates to compact hash keys, which are then employed to index a set of optimized features for various vision signal reconstruction. \cite{xie2023diner} propose disorder-invariant INR by incorporating a hash-table to a traditional INR backbone. This innovative representation ensures that coordinates map into the same distribution, which is turn enables the projected signal to be more accurately modeled by the subsequent INR network. As a result, this method significantly mitigates the issue of spectral bias, leading to improved performance in various signal processing tasks.

By expanding and refining the potential of hash-based INRs through the incorporation of learned hash functions, these methodologies play a pivotal role in propelling the continuous evolution of neural scene representation. Nevertheless, an important consideration arises due to the fact that hash-based INRs employ hash-keys as the network input instead of the raw coordinates (see Sec.~\ref{sec:INR_performance} for details). This approach could potentially lead to a substantial reduction or even complete loss of their regularization capacity when handling unseen coordinates. The consequence of this diminished regularization is the emergence of disruptive noisy artifacts, particularly in tasks that necessitate interpolation. Hence, it becomes imperative to strike a harmonious balance between the augmented expressive power offered by hash-based INRs and the preservation of robust regularization capabilities.

\subsection{Regularizing the INR}
Apart from expressive power, regularization (\textit{i.e.}, interpolations on points not present in the training set) is another crucial metric that determines the performance of INRs for signals requiring continuous representation, such as neural radiance field optimization for high-quality novel view synthesis at any position~\citep{mildenhall2020nerf}.

Although numerous efforts have been made to regularize classical deep neural networks~\citep{srivastava2014dropout,maennel2018gradient,heiss2019implicit,kubo2019implicit}, there are a few works on regularizing the INR. \cite{ramasinghe2022frequency} find that a shallow MLP network tends to suppress lower frequencies when the hyper-parameters or depth is increased, resulting in noisy interpolations. They propose the gradient loss built from the second to last layer of the MLP for regularizing the noise. \cite{li2023regularize} analyse the poor interpolations of INR on nonuniform sampling following the neural tangent kernel theory~\citep{jacot2018neural}, they propose the Dirichlet energy regularization by measuring the similarities between rows/columns for a 2D image. 


However, all existing works concentrate on regularization for INRs with continuous and analytical gradient flow between the input coordinate and network output. As analyzed in Sec.~\ref{sec:INR_performance}, the continuous function modeled in hash-based INRs is built upon hash-keys rather than input coordinates, breaking the gradient flow between the input coordinate and hash-key and resulting in noisy interpolations in hash-based methods. Consequently, previous regularization methods cannot be directly applied to hash-based INRs.

\section{Behaviors of the Hash-based INRs}
\label{sec:INR_performance}
\subsection{Principle of the Hash-based INRs}
Given a paired signal $Y=\{(\vec{x}_i,\vec{y}_i)\}_{i=1}^{N}$ with length $N$, where $\vec{x}_i$ and $\vec{y}_i$ are the $i$-th $d_{in}$-dimensional coordinate and the corresponding $d_{out}$-dimensional attribute. For the convenience of derivation, $d_{in}$ is set as $1$, thus the term $\vec{x}_i$ is written as ${x}_i$ in the following sections. INR characterizes the paired signal by modeling a function mapping between the input coordinate ${x}_i$ and the output attribute $\vec{y}_i$ using an MLP, that is
\begin{equation}
\begin{aligned}
    \mathbf{z}^{0}&=\gamma ({x}_i), \\
    \mathbf{z}^{j}&=\rho^{j}(\mathbf{W}^{j}\mathbf{z}^{j-1}+\mathbf{b}^{j}),\: j=1,...,J-1\\
    f_{\theta}({x})&=\mathbf{W}^{J}\mathbf{z}^{J-1}+\mathbf{b}^{J}
\end{aligned},
\label{eqn:MLP_structure}
\end{equation}
where $\gamma(\cdot)$ refers to the continuous preprocess encoding function such as the commonly used Fourier encoding~\citep{tancik2020fourier}, $\mathbf{z}^{j}$ refers to the output of the $j$-th layer of the network, $\theta=\{\mathbf{W}^{j}, \mathbf{b}^{j}\}_{j=0}^{J}$ are the network parameters, where the $\mathbf{W}^{j}, \mathbf{b}^{j}$ are the weight and bias of the $j$-th network layer.

However, there is a spectral bias in the function modeled using the Eqn.~\ref{eqn:MLP_structure} that the $f_{\theta}(x)$ is composed of a linear combination of certain harmonics of the $\gamma(\cdot)$~\citep{yuce2022structured}, \textit{i.e.},
\begin{equation}
\begin{aligned}
      &f_{\theta}({x})\in \{\sum_{\omega'\in \mathcal{F}_{\omega}}c_{\omega'}\sin (\langle\omega',x\rangle+\phi_{\omega'})\}\\
      &c_{\omega'},\phi_{\omega'}\in \mathbf{Q},  
\end{aligned}
\label{eqn:spectral_bias}
\end{equation}
where $\mathcal{F}_{\omega'}$ is the frequency set determined by the encoded frequencies in $\gamma(\cdot)$, $\mathbf{Q}$ is the set of rational number.

\begin{figure*}
\centering
\includegraphics[width=\textwidth]{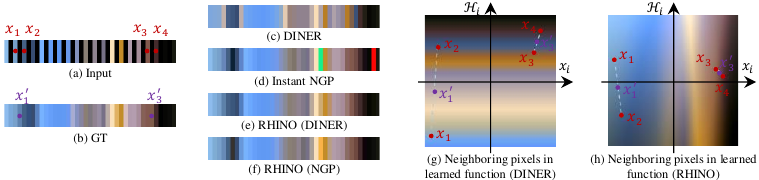}
\caption{Comparisons of the hash-based INRs and the proposed RHINO. (a) is the input where the black bands refer to the points to be interpolated. (b) the ground truth. (c) and (d) are interpolations by applying the hash-based INRs, \textit{i.e.}, the DINER~\citep{xie2023diner} and the Instant NGP~\citep{muller2022instant} directly. (e) and (f) are the results of that using the proposed regularization. (g) and (h) visualize the neighboring pixels and interpolations in the actually learned function of DINER and the RHINO with the DINER backbone, respectively.}
\label{fig:reg_dim1_cmp}
\end{figure*}

To overcome this limitation, the hash-table $\mathcal{H}$~\citep{muller2022instant,kang2022pixel,xie2023diner} is introduced to replace the role of the continuous preprocess function $\gamma(\cdot)$, that
\begin{equation}
    \mathbf{z}^{0}= \mathcal{H}({x}_i),
\label{eqn:MLP_structure_hash}
\end{equation}
where the keys in the hash-table $\mathcal{H}$ are also updated during the network training following the chain rule. Note that, because the hash-key is indexed according to the relative position of the ${x}$ instead of transforming ${x}$ with continuous functions, the chain rule is broken between the input coordinate ${x}$ and the hash-key, as a result, $\mathcal{H}(x_i)$ could be directly written as $\mathcal{H}_i$. Since the hash-key is also updated in the training process, the optimization of the hash-table in INR could be viewed as finding the solution $\mathcal{H}_i$ of the function $f_{\theta}(\mathcal{H}_i)=\vec{y}_i$ instead of fitting a function $\theta$ with known input $\mathcal{H}_i$ and output $\vec{y}_i$. As a result, the hash-based INRs~\citep{muller2022instant,xie2023diner} have much stronger expressive power than traditional function-expansion-based INRs~\citep{tancik2020fourier,sitzmann2020implicit}.

\subsection{Hash-table Turns off the Regularization}
When applying these two types of INRs (\textit{i.e.}, function-expansion-based and hash-based) to the inverse problems with forward physical process $\mathcal{P}$, the optimizations of them are,
\begin{subequations}
\label{eqn:loss_func_cmp}
\begin{align}
    \theta ^*
    &=\argmin _{\theta}\mathcal{L}\left(
    \mathcal{P}\left(\{ f_{\theta}( x_i ) \}_{1}^{N}\right),
    \mathcal{P}\left(\{ \vec{y}_i\}_{1}^{N}\right) \right) \label{eqn:loss_func_cmp:CON}\\
    \theta ^*,\mathcal{H}^*
    &=\argmin _{\theta ,\mathcal{H}}\mathcal{L}\left(
    \mathcal{P}\left(\{ f_{\theta}( \mathcal{H}_i ) \}_{1}^{N}\right),
    \mathcal{P}\left(\{ \vec{y}_i\}_{1}^{N}\right) \right) ,
    \label{eqn:loss_func_cmp:hash}
\end{align}
\end{subequations}
 where $\mathcal{H}_i$ is the corresponding hash-key of the $i$-th coordinate ${x}_i$. It is noticed that the length of $\mathcal{H}$ is not always equal to the length $N$ of the signal $Y$, actually it could either be larger or smaller than $N$ according to different strategies of hash indexing, \textit{e.g.}, $|\mathcal{H}|=N$ when the full-resolution hash-table is used~\citep{xie2023diner}, $|\mathcal{H}|<N$ when multi-scale hash-tables are used for representing gigapixel images~\citep{muller2022instant}. Mostly, $\mathcal{H}_i$ is represented by a linear weighting summation~\citep{xie2023diner,kang2022pixel,abou2022particlenerf} or concatenation~\citep{muller2022instant} of the elements $h_i$ in $\mathcal{H}$, \textit{i.e.},
 \begin{equation}
 \begin{aligned}
     \mathcal{H}_i\in
     &\left\{\sum_{j=1}^{|\mathcal{H}|}a_{ij}h_{j}\right\}\cup  
\left\{\bigoplus_{j=1}^{|\mathcal{H}|}a_{ij}h_{j}\right\} \\
     h_{j}\in &\mathcal{H},a_{ij}\in\mathbf{Q}^{+}\cup\{0\}
 \end{aligned}
 \end{equation}
where $\mathbf{Q}^{+}$ is the set of positive rational number.


The traditional INR (Eqn.~\ref{eqn:loss_func_cmp:CON}) directly learns the continuous function between the ${x}_i$ and the $\vec{y}_i$, which means that the interpolation on unsampled points (\textit{e.g.}, ${x}_j$, $j\notin \{1,2,...,N\}$) also follows the regularization of the spectral bias~\citep{yuce2022structured}. On the other hand,
because the gradients flow between the ${x}_i$ and the $\vec{y}_i$ is broken, the hash-based INR (Eqn.~\ref{eqn:loss_func_cmp:hash}) actually learns the continuous function between the $\mathcal{H}_i^*$ and the $\vec{y}_i$. Following the spectral bias~\citep{yuce2022structured}, there is a regularization on the hash-key ${H}_j^*$ that does not appear in the training process. Unfortunately, because the mapping function between the coordinate ${x}_i$ and the hash-key $\mathcal{H}_i^*$ is discontinuous and not analytical, the interpolation on unsampled point ${x}_j$ is also discontinuous and not analytical, resulting in the lost regularization.

Fig.~\ref{fig:reg_dim1_cmp} demonstrates the above observation. Given a 1D signal with $d_{in}=1$ and $d_{out}=3$, hash-based INRs (DINER \citep{xie2023diner} and Instant NGP~\citep{muller2022instant}) provide high accuracy for representing the coordinates appeared in the training set (Fig.~\ref{fig:reg_dim1_cmp} (c) and (d)). However, both of them produce unreliable results for interpolated coordinates (\textit{i.e.}, the black points in Fig.~\ref{fig:reg_dim1_cmp}(a)). Fig.~\ref{fig:reg_dim1_cmp}(g) visualizes the learned function $f_{\theta}(\mathcal{H}_i)$ of the neural network in DINER. It is noticed that the neighboring pixels (\textit{e.g.}, $x_1$ and $x_2$, or $x_3$ and $x_4$) become far from each other, and the line between them will pass through several color bands, thus unreliable interpolations (Fig.~\ref{fig:reg_dim1_cmp} (c) and (d)) are produced. In summary:
\begin{prop}
\label{prop:hash_no_regu}
The introduction of the hash-table significantly enhances the expressive power of the INRs at the cost of the lost regularization for interpolations.
\end{prop}

\begin{figure*}
\centering
\includegraphics[width=0.85\textwidth]{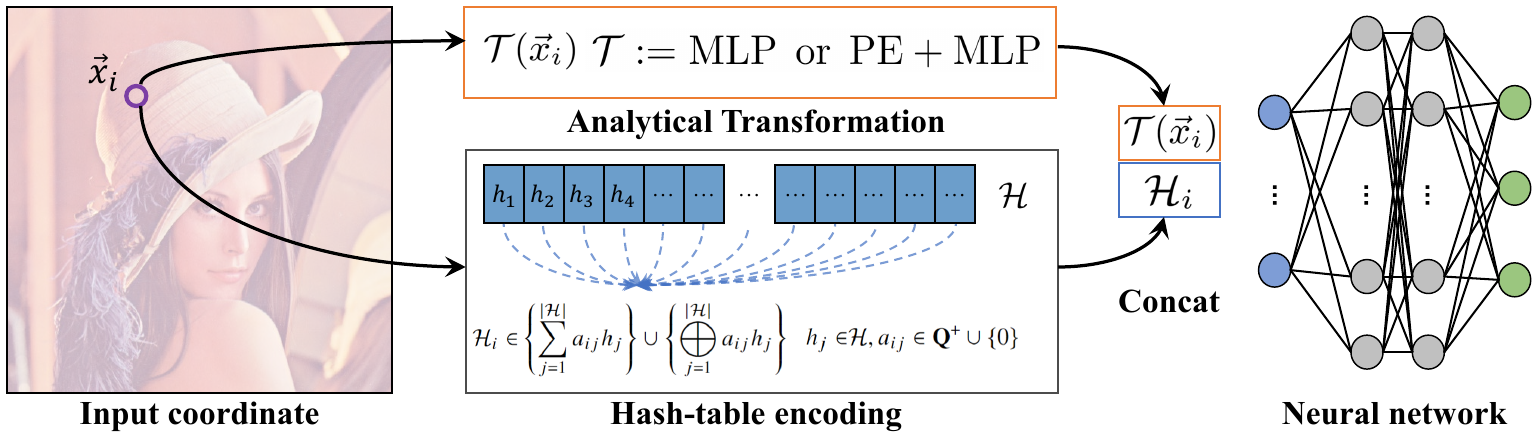} 
\caption{The architecture of the proposed RHINO. The regularization of previous hash-based INR could all be improved by introducing the input coordinate with an analytical and continuous transformation additionally.}
\label{fig:network_architecture}
\end{figure*}

\section{Regularizing the Hash-based INR}
\label{sec:INR_regularization}
\subsection{Rebuilding the connections}
Because the lost regularization is caused by the broken gradients flow between the ${x}_i$ and the $\vec{y}_i$, it is useless to compensate the regularization by introducing the traditional smoothness terms (\textit{e.g.}, the total variation~\citep{rudin1992nonlinear} and the low rank~\citep{hu2021low} priors) in the loss function. In the hash-based INR, the key for compensating the lost regularization is adding the input coordinate ${x}$ with an analytical and continuous transformation $\mathcal{T}$ to the network, \textit{i.e.},
\begin{equation}
\begin{aligned}
    &\theta ^*,\mathcal{H}^*\\
    =&\argmin _{\theta ,\mathcal{H}}\mathcal{L}\left(
    \mathcal{P}\left(\{ f_{\theta}( \mathcal{H}_i,\mathcal{T}( x_i ) ) \}_{1}^{N}\right),
    \mathcal{P}\left(\{ \vec{y}_i\}_{1}^{N}\right) \right) .
\end{aligned}
\end{equation}



Following this idea, we propose the RHINO as shown in Fig.~\ref{fig:network_architecture}, where the architecture of previous hash-based INRs is little altered with an additional input $\mathcal{T}(x_i)$. Note that, because the proposed regularization is a universal framework for various encoding (\textit{e.g.}, single scale full-resolution hash-table encoding~\citep{xie2023diner} and multi-scale hash-table encoding~\citep{muller2022instant}) methods using the hash-table, the middle-bottom panel in Fig.~\ref{fig:network_architecture} only provides the sketch for constructing hash-keys $\mathcal{H}_i$. There are several candidates for the selection of the transformation $\mathcal{T}$, such as the identical transformation, a standard MLP network or an MLP with positional encoding. As we will show in the next subsection, these candidates could all improve the regularization of previous pure hash-based methods, but with different behaviors according to the encoded frequencies.

\subsection{Analysis of Regularization}
For better visualizing the structure of the regularization, we focus on the case that the width of the hash-key $\mathcal{H}_i$ is $1$ and the linear weighting summation is used to produce the indexed hash-key. In this case, the learned INR actually builds a continuous function with 2D input, \textit{i.e.}, the $x_i$ and $\mathcal{H}_i$.

Because it is difficult to analyse the properties of the $f_{\theta}(\mathcal{H}_i,\mathcal{T}(x_i))$ directly, the 'slicing strategy' is firstly used for analysing different dimensions of the $f_{\theta}(\mathcal{H}_i,\mathcal{T}(x_i))$ one by one, then the global property is summarized. In detail, by fixing the $\mathcal{T}(x_i)$ with different real values, respectively, \textit{e.g.}, $\mathcal{T}(x_i)=0,...,1$, the 2D function $f_{\theta}(\mathcal{H}_i,\mathcal{T}(x_i))$ is reduced to a 1D function (\textit{i.e.}, $f_{\theta}(\mathcal{H}_i,0),...,f_{\theta}(\mathcal{H}_i,1)$). The property of each 1D function is equivalent to the previous pure hash-based methods that, 1) the connection between the coordinate and network output is also broken and 2) the distribution of each 1D function also follows the spectral bias. In other words, the regularization is still lost in each 1D function. 

However, things change when the attention is focused on another input dimension $\mathcal{T}(x_i)$ by fixing the $\mathcal{H}_i$ with real values, \textit{i.e.}, $f_{\theta}(0,\mathcal{T}(x_i)),...,f_{\theta}(1,\mathcal{T}(x_i))$. Because there is an analytical connection between the input coordinate $x_i$ and network output, these 1D functions are all continuous functions regarding the $x_i$ and the distribution of each 1D function here follows the spectral bias. In other words, the interpolation on unsampled point $x_j$ also follows the spectral bias.

\begin{figure}
\centering
\includegraphics[width=\linewidth]{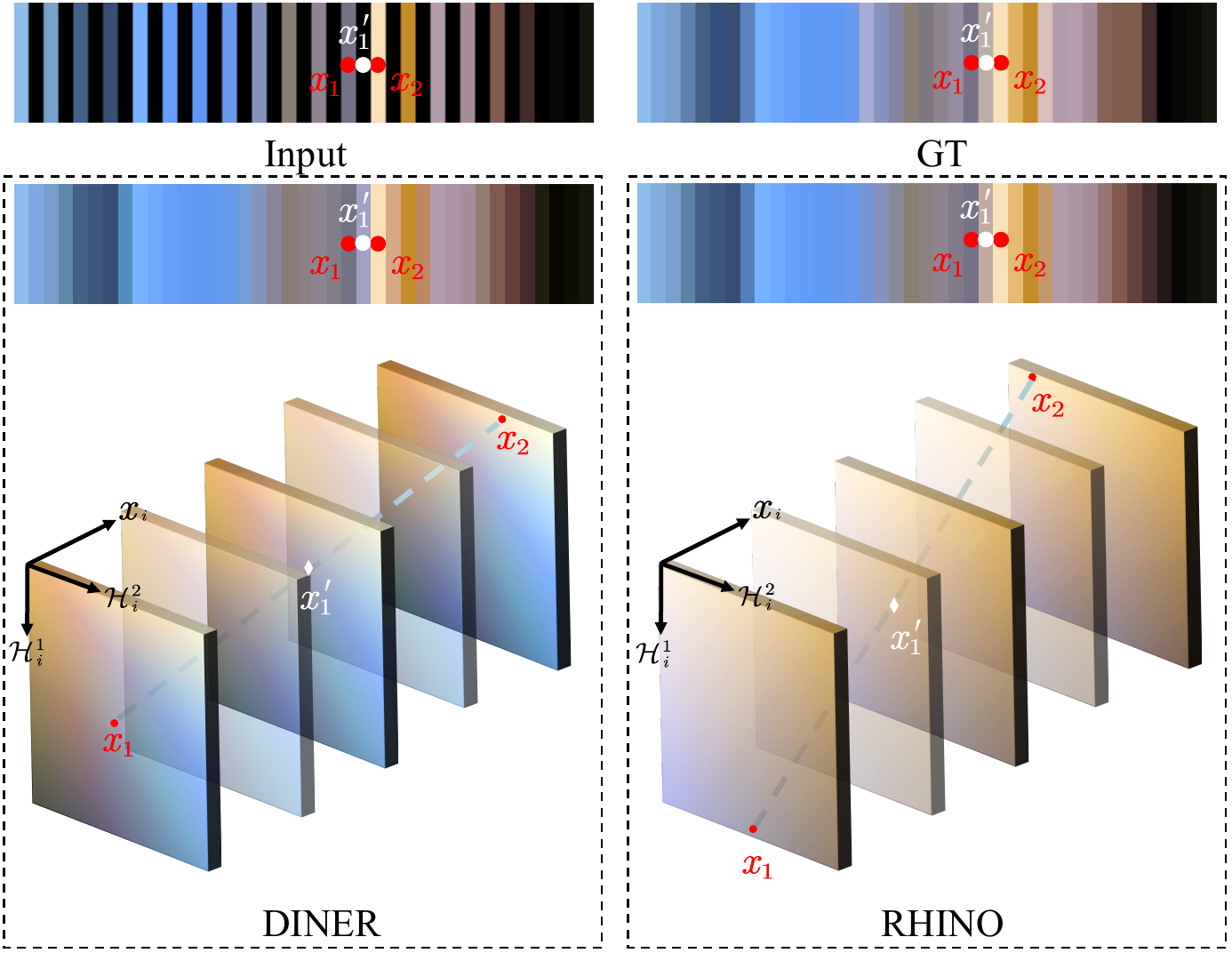}
\caption{Comparison of the learned functions with and without $\mathcal{T}$ for the hash-key $\mathcal{H}_i$ with width $2$. RHINO corrects the wrong purple interpolation $x_1'$ with the right khaki color.}
\label{fig:reg_dim2_cmp}
\end{figure}

Fig.~\ref{fig:reg_dim1_cmp} visualizes the above process and the comparisons. Compared with previous pure hash-based INRs (Fig.~\ref{fig:reg_dim1_cmp}(c) and (d)), the proposed regularization significantly improves their performance on interpolated points (Fig.~\ref{fig:reg_dim1_cmp}(e) and (f)). Fig.~\ref{fig:reg_dim1_cmp}(g) and (h) visualize the learned functions without and with regularization. Compared with the one without regularization where the line segment between neighboring pixels will pass through several color bands, the neighboring pixels in the one with regularization are stayed in a similar neighboring color band where the intensity change is more smooth, thus more reliable interpolations (Fig.~\ref{fig:reg_dim1_cmp}(e) and (f)) are produced. Note that, the regularization performance of the RHINO is not limited to the hash-key $\mathcal{H}_i$ with width $1$, it also works for the case that the width is larger than $1$. Fig.~\ref{fig:reg_dim2_cmp} shows the learned function with the width $2$, it could be noticed that the lost regularization along the $x_i$-axis is compensated by adding an analytical function $\mathcal{T}$, resulting the fact that the wrong purple point $x_1'$ in the DINER is corrected as a khaki point in the RHINO.

\begin{figure}
\centering
\includegraphics[width=\linewidth]{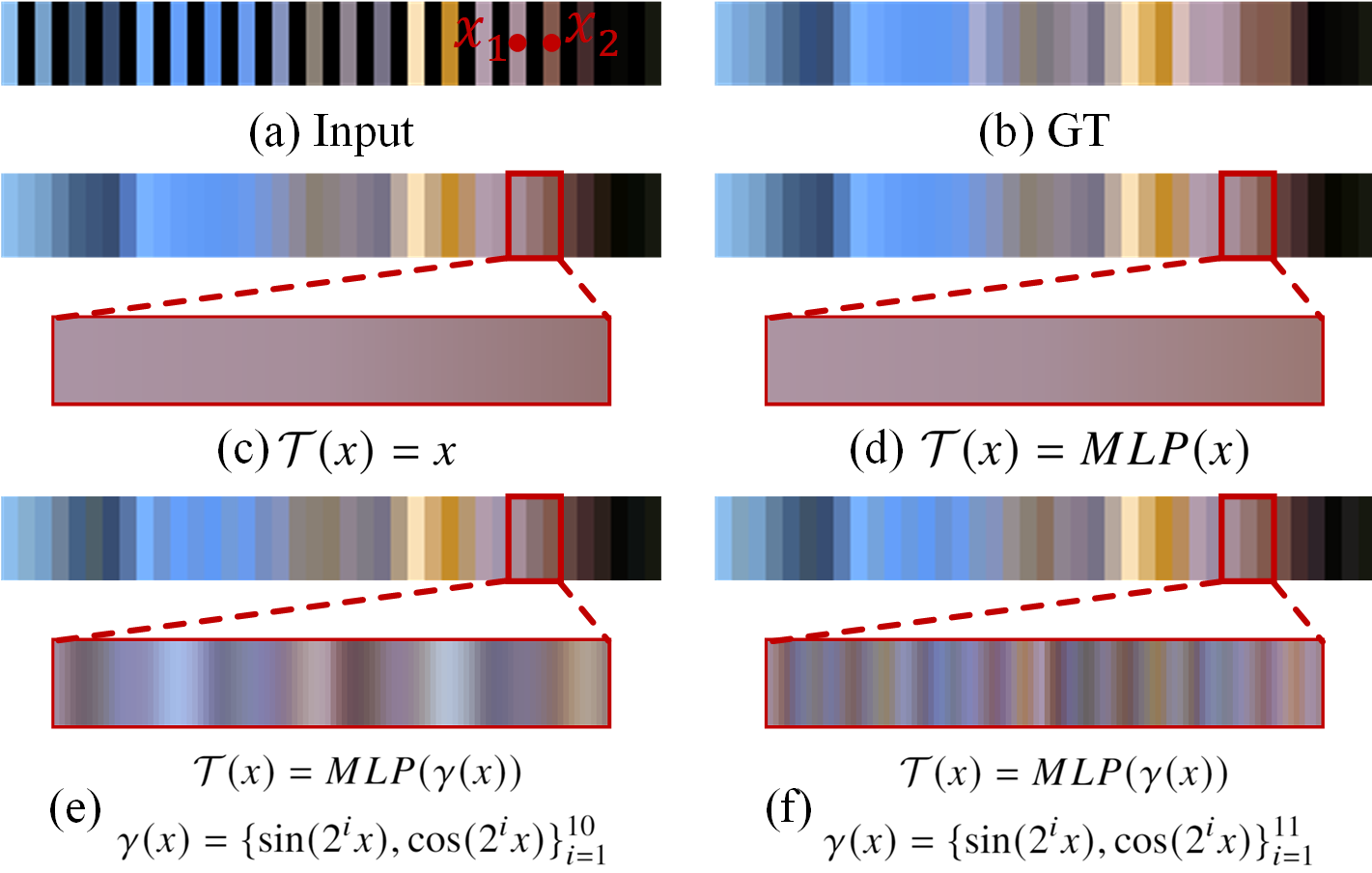} 
\caption{Comparisons of the proposed regularization with different analytical function $\mathcal{T}$. Different analytical functions $\mathcal{T}$ have similar behavior when only one point is interpolated, however the interpolated textures on dense samples also follow the spectral bias (\textit{i.e.}, the Eqn.~\ref{eqn:spectral_bias}).}
\label{fig:reg_dim1_cmp:dif_T}
\end{figure}

\noindent \textbf{Analytical Transformation $\mathcal{T}$}. Different analytical transformation function $\mathcal{T}$ have different regularization behaviors. Fig.~\ref{fig:reg_dim1_cmp:dif_T} compares the interpolations with different analytical functions, \textit{i.e.}, the identical function $\mathcal{T}(x)=x$, a standard MLP without encoding, and Fourier-encoded MLPs~\citep{tancik2020fourier} with different encoding frequencies. It is noticed that all 4 results produce reliable interpolations when only one point is queried between neighboring points, however the texture distribution changes when denser points are queried. To elaborate, low-frequency textures are generated when employing the identical function as well as the standard MLP without encoding. Conversely, higher-frequency interpolations emerge when the high-frequency components are encoded within the preprocessing function $\gamma(\cdot)$ of $\mathcal{T}(\cdot)$. In summary:
\begin{prop}
The additional connection established between the input coordinates and the network enhances the regularization of hash-based INR, where the regularization behavior are influenced by the encoded frequencies within the added analytical function associated with the input coordinates.
\label{prop:hash_dif_T}
\end{prop}

\section{Experiment}
To better demonstrate the performance of the proposed regularization under different inverse problems (\textit{i.e.}, representation, regularization and different dimensions), RHINO is applied in four separate tasks including the 2D image representation, 3D signed distance function representation, 5D static and 6D dynamic neural radiance field reconstruction. Note that, the analytical function $\mathcal{T}(x)$ is set as the standard MLP with one hidden layer ($1\times 64$ neurons), same output dimensions (as the input coordinate) and position encoding ($\gamma(x)=\{\sin(2^{i}\pi x), \cos(2^{i}\pi x)\}_{i=0}^{9}$) in all following experiments.

\begin{figure*}[!]
\centering
\includegraphics[width=\textwidth]{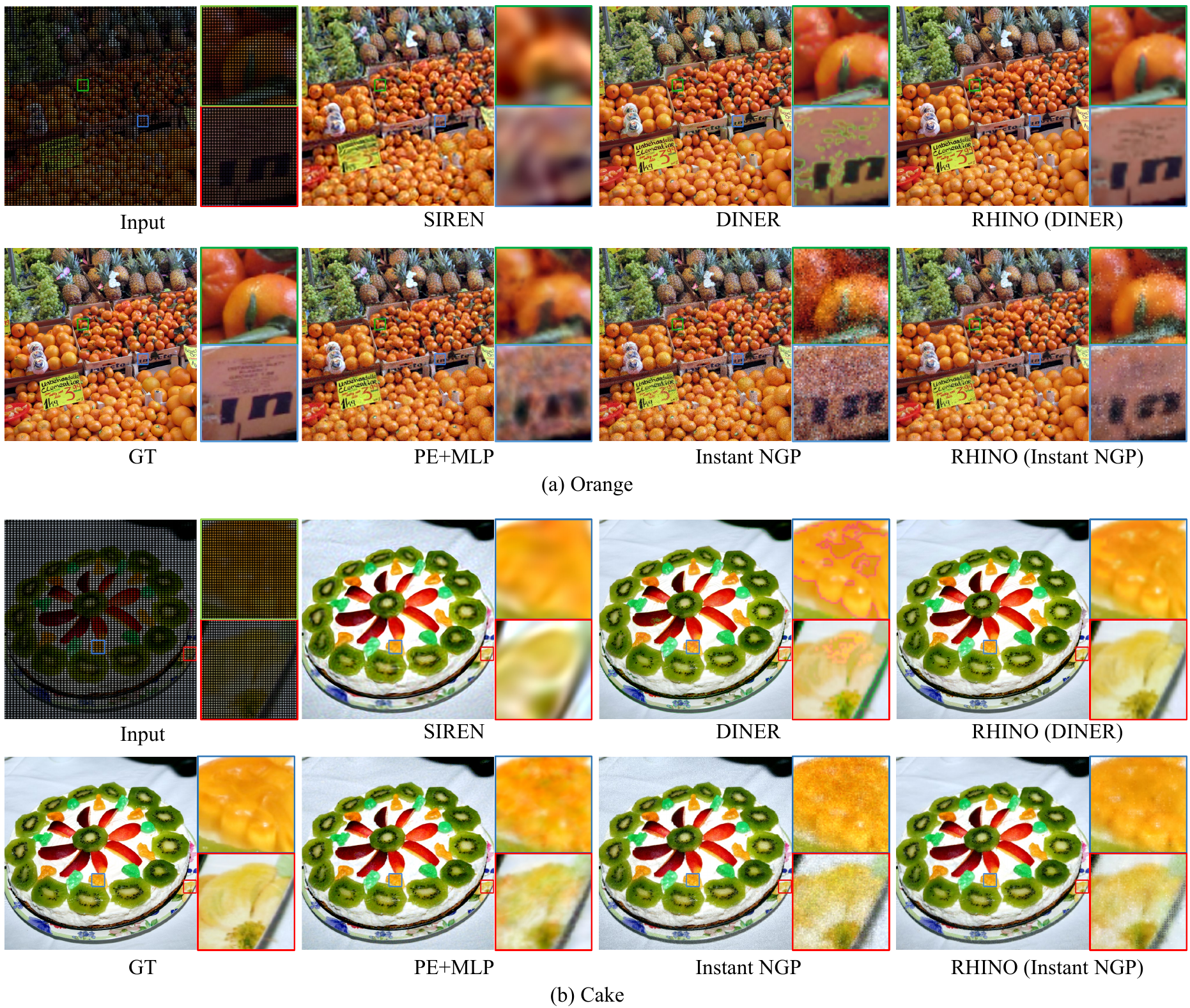} 
\caption{Comparisons of RHINO with different INR backbones. }
\label{fig:res_img_fitting_quality}
\end{figure*}

\subsection{Image Representation}
We use an image representation task to verify the performance of the proposed RHINO and to demonstrate its regularization behavior explicitly. In the experiment, the total 100 high-resolution ($1200\times 1200$) images from the SAMPLING category of the TESTIMAGES dataset~\citep{asuni2014testimages} are used for evaluation. In the experiment, $1/4$ pixels are uniformly sampled for training different INRs, then all pixels are used for evaluated the regularization behavior. A total of four baselines are compared, including two classical function-expansion-based methods, PE+MLP~\citep{tancik2020fourier} and SIREN~\citep{sitzmann2020implicit}, as well as two SOTA hash-based methods, Instant NGP~\citep{muller2022instant} and DINER~\citep{xie2023diner}. As the proposed RHINO is a universal framework, the results of RHINO on the DINER and the Instant NGP are both compared. Note that, the Torch-NGP implementation by~\cite{torch-ngp} is used throughout the experiment since it is easier to be modified compared with the original Tiny-cuda-nn version~\citep{muller2022tinycudann}. For a fair comparison, all methods are set with the same network configuration, \textit{i.e.}, 2 hidden layers with 64 neurons per layer, loss function with $L_2$ distance between the network output and the ground truth, and are trained for 3000 iterations using the Pytorch and the Adam optimizer with a batchsize $600\times 600$ on an Nvidia A100 40GB GPU. 

\begin{table}[!t]
  \centering
  \caption{Comparisons of training time (sec.) for fitting a 2D image with 3000 epochs.}
  \label{tab:time_image}
  \begin{tabular}{@{}lcccccc@{}}
    \toprule
     & RHINO & RHINO & \multirow{2}{*}{DINER} & \multirow{2}{*}{NGP} & PE & \multirow{2}{*}{SIREN}\\
     & (DINER) & (NGP) & & &+MLP &\\
    \midrule
    Time& 15.24 & 15.37 & 11.51 & 9.44 &14.50 &14.73 \\
    \bottomrule
  \end{tabular}
\end{table}

\begin{figure}[!t]
\centering
\includegraphics[width=0.82\linewidth]{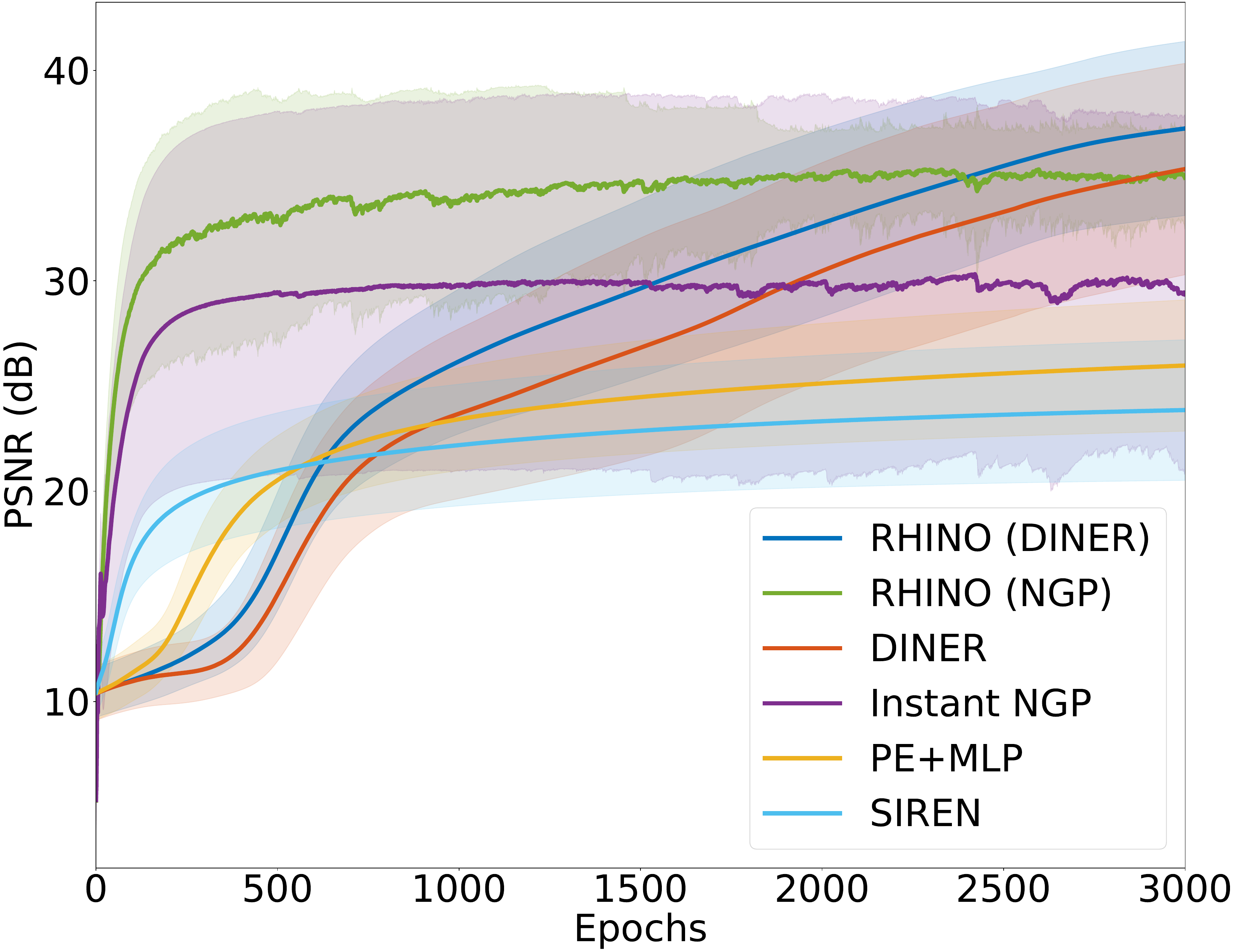} 
\caption{Training curves of different INRs on fitting 2D image. The main curve and transparent regions are the mean and standard deviation across different images, respectively.}
\label{fig:res_img_psnr_curve}
\vspace{-0.6cm}
\end{figure}

\definecolor{red}{rgb}{1.0000,0.5686,0.5059}
\definecolor{orange}{rgb}{1.0000,0.7373,0.5059}
\definecolor{yellow}{rgb}{1.0000,0.8431,0.5059}

\begin{table*}[!t]
\centering
\begin{tabularx}{0.95\textwidth}{@{}llCCCCC@{}}
\toprule
 & Scene  & Armadillo & Dragon & Lucy & Thai Statue  & Avg \\
\midrule
 \multirow{5}{*}{IOU $\uparrow$}  & RHINO (DINER) &\cellcolor{red}0.990 &\cellcolor{red} 0.985 &  0.966 & 0.973 & 0.978\\
                        & RHINO (NGP) & \cellcolor{red}0.990 &\cellcolor{red} 0.985 & \cellcolor{orange} 0.968 & \cellcolor{red} 0.980 &\cellcolor{red}0.981\\
                        & Bacon &\cellcolor{red}0.990 & 0.981 &  \cellcolor{red}0.969 & \cellcolor{orange}0.979 & \cellcolor{orange}0.980\\
                        & DINER &\cellcolor{orange}0.976 & 0.978 &  0.955 & 0.971 & 0.970\\
                        &Instant NGP&\cellcolor{red}0.990 &\cellcolor{orange}0.983 & 0.967 &\cellcolor{red}0.980 & \cellcolor{orange}0.980\\
\midrule                        
 \multirow{5}{*}{Chamfer $\downarrow$}  & RHINO (DINER) &3.230e-6 & 1.891e-6 &  3.011e-6 & 1.927e-6 & 2.515e-6\\
                        & RHINO (NGP) & \cellcolor{red}3.215e-6 & \cellcolor{red}1.853e-6 &  \cellcolor{orange}2.909e-6 & \cellcolor{red}1.898e-6 & \cellcolor{red}2.469e-6\\
                        & Bacon &3.243e-6 & 1.895e-6 &  3.052e-6 & 1.945e-6 & 2.533e-6\\
                        & DINER &2.766e-5 & 8.758e-6 &  1.662e-5 & 1.076e-5 & 1.595e-5\\
                        &Instant NGP&\cellcolor{orange}3.228e-6 & \cellcolor{orange}1.859e-6 &  \cellcolor{red}2.899e-6 & \cellcolor{orange}1.904e-6 & \cellcolor{orange}2.475e-6\\    
\midrule                        
 \multirow{5}{*}{Time (min.)$\downarrow$} & RHINO (DINER) & 62.48 & 56.15 &  148.93 & 344.88 & 153.11\\
                        & RHINO (NGP) & \cellcolor{orange}0.60 & \cellcolor{orange}0.73 & \cellcolor{red}0.98& \cellcolor{orange}1.73 & \cellcolor{orange}1.01 \\
                        & Bacon & 82.10 & 77.60 & 163.90 & 334.92 & 164.63 \\
                        & DINER & 55.27 & 49.23 & \cellcolor{orange}132.40 & 324.47 & 140.34\\
                        &Instant NGP& \cellcolor{red}0.57 & \cellcolor{red}0.72 & \cellcolor{red}0.98 & \cellcolor{red}1.72 & \cellcolor{red}1.00 \\                       
\bottomrule
\end{tabularx}
\caption{Quantitative comparison on 3D shape representation.  We color code each cell as \colorbox{red}{best}, \colorbox{orange}{second best}.}
\label{tab:res_sdf}
\end{table*}

Figs.~\ref{fig:res_img_first} and~\ref{fig:res_img_fitting_quality} compare the regularization behaviors of different methods at 3000 epochs qualitatively. Because there is an analytical connection between the input coordinate and the output attribute, there are no obvious artifacts in the function-expansion-based INRs (\textit{i.e.}, the SIREN and the PE+MLP). However, as analysed in Sec.~\ref{sec:INR_regularization} that the hash-based INRs focus on optimizing the mapping between the signal attribute and the hash-key instead of the coordinate-self, undesired artifacts appear when the interpolation is applied to the last $3/4$ pixels, such as the nose and the pineapple in Fig.~\ref{fig:res_img_first}, the orange and the box in the Orange as well as the corn in the Cake in Fig.~\ref{fig:res_img_fitting_quality}. RHINO rebuilds the analytical connection between the input coordinate and the output attribute, as a result, the noisy artifacts could be significantly alleviated when the RHINO is applied to these hash-based INRs.

Fig.~\ref{fig:res_img_psnr_curve} shows the PSNR of various methods for fitting the sampled $1/4$ pixels at different epochs. Hash-based methods (DINER and Instant NGP) outperform the function-expansion-based methods (SIREN and PE+MLP). RHINO further improves the accuracy of these hash-based methods while the trend of the curves are also similar with their original versions. Additionally, it is noticed that RHINO (NGP) provides  more stable training behavior compared with the original Instant NGP (\textit{e.g.}, the green curve has a smaller  fluctuating range than the purple curve). Tab.~\ref{tab:time_image} lists the training time of 3000 epochs. RHINO is slightly slower than the function-expansion-based INRs.

\subsection{3D Shape Representation}

In this section, we demonstrate the representation capabilities of RHINO for representing 3D shapes as SDF, which measures the distance $s$ between the given spatial point $\vec{x}$ and the closest surface as a continuous function. The sign of the distance indicates whether the point is inside (negative) or outside (positive) the watertight surface, 
\begin{equation}
\mathrm{SDF}(\vec{x})=s: \vec{x} \in \mathbb{R}^3, s \in \mathbb{R}.
\end{equation}
To visualize this implicit surface, it can be rendered using raycasting or rasterization techniques on a mesh generated by algorithms such as Marching Cubes ~\citep{lorensen1998marching}. This method allows us to convert the representation of the shape into a more tangible and visually interpretable form, enabling visualization and analysis of the 3D surface. 

In the experiment, four shapes from the \cite{standord-3D-scanning} are used for evaluation, namely the Armadillo, Dragon, Lucy, and Thai Statue. In the experiment, $10$k points are randomly sampled in each iteration during the training process (the same points are used in all methods), then a 512 cubed grid is extracted for evaluation and visualization. Three baselines are compared, \textit{i.e.}, Bacon~\citep{lindell2022bacon}, DINER~\citep{xie2023diner} and Instant NGP~\citep{muller2022instant}. The performance of the RHINO is verified by applying it to the latter two hash-based methods. 


Tab.~\ref{tab:res_sdf} provides a quantitative comparison of the performance of various methods. RHINO demonstrates significant improvements in both DINER and Instant NGP across almost all four shapes. Furthermore, the increase in computational time for RHINO compared to their original versions is minimal (e.g., $153.11$ vs. $140.34$ and $1.01$ vs. $1.00$). It's worth noting that Instant NGP and its RHINO version require considerably less time than other methods. This can be attributed to the rapid convergence of Instant NGP, which achieves convergence in just $2,000$ iterations compared to $200,000$ for other methods. Additionally, it's noteworthy that the relative increase in time cost (approximately $9\%$ from $(153-140)/140$) in this context is much smaller than that observed in image fitting (approximately $36\%$ from $(15-11)/11$). This disparity arises because the construction of the loss function contributes more to computational cost in SDF fitting tasks compared to running the network.

\begin{figure*}[!]
\centering
\includegraphics[width=\textwidth]{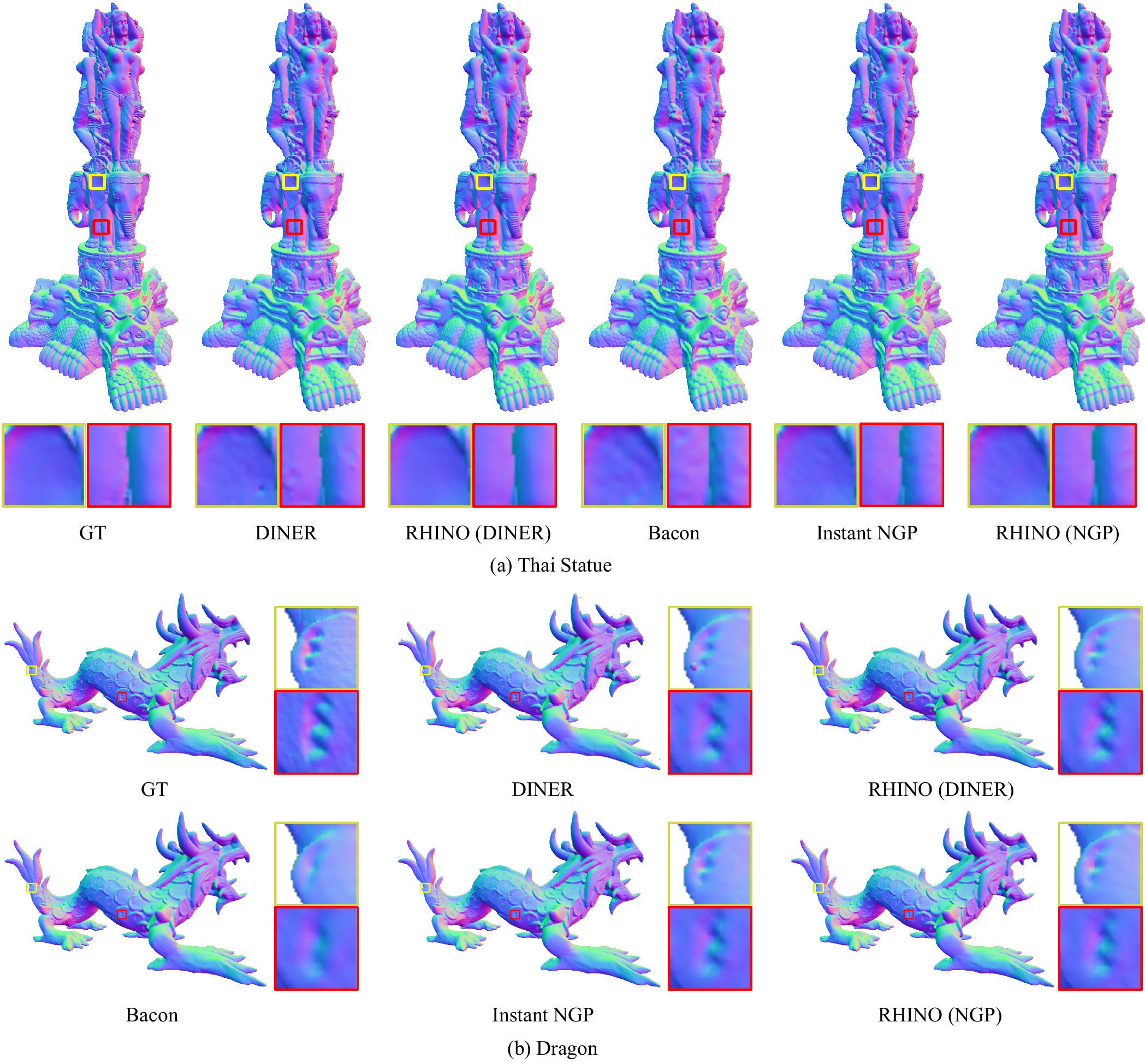}
\caption{Shape fitting results. The results indicate that our approach outperforms other methods in representing 3D shapes by effectively capturing smoothness, while other methods exhibit noise artifacts.}
\label{fig:res_sdf}
\end{figure*}

Fig.~\ref{fig:res_sdf} visualizes the difference in reconstructed details. Because the learned function takes the hash-key as an independent variable instead of the coordinate-self, DINER produces undesired artifacts in interpolated areas such as the bulge in the elephant ear and leg in Thai Statue and the squama in Dragon. RHINO (DINER) overcomes this problem and produces more smooth results. Instant NGP alleviates the problem by introducing multi-scale hash grids, however, similar artifacts also appear in the gap between the elephant legs (zoom-in red box in Thai Statue of Fig.~\ref{fig:res_sdf}. RHINO (NGP) also improves Instant NGP with more clear reconstructed legs. Bacon focuses on reconstructing the SDF signal following a scale-by-scale behavior using the $\sin$ base function, therefore the unpopular harmonics wave appears in the reconstructed elephant ear and leg which should have smooth surfaces.

In contrast to the task of image fitting, the enhancements achieved by RHINO in SDF representation appear relatively modest. This can be ascribed to the distinct sampling strategies employed. In image fitting, uniform samplings span the entire domain, whereas in SDF representation, only points proximate to the surface are sampled. Because the points requiring interpolation in SDF representation are closer to the trained points compared to the situation in image fitting, the relative improvement in SDF representation is less pronounced than what is observed in image fitting.

\begin{table*}[!t]
\centering
\begin{tabularx}{1\textwidth}{@{}ClCCCcCCCCC@{}}
\toprule
& Scene  & Materials & Hotdog &  Ship &Mic  & Drums & Lego & Chair & Ficus& Avg\\
\midrule
\multirow{7}{*}{PSNR$\uparrow$}  & RHINO (DVGO) & \cellcolor{red}30.36& \cellcolor{red}38.69&\cellcolor{orange}31.90 &\cellcolor{yellow}36.15 & \cellcolor{red}26.61 & \cellcolor{red}36.27&\cellcolor{red}37.43 &\cellcolor{yellow}32.71 &\cellcolor{red}33.77 \\
& RHINO (NGP) &29.49  & \cellcolor{yellow}38.29  & \cellcolor{red}32.05  & \cellcolor{red}36.68  & \cellcolor{yellow}26.33  & \cellcolor{yellow}35.86  & \cellcolor{yellow}36.89  & \cellcolor{red}33.25  & \cellcolor{orange}33.61 \\
& NeRF &29.39 & 36.73& 29.17& 33.08& 25.61& 31.54& 33.90& 28.94& 31.04\\
& Plenoxels & 29.18& 36.03& 29.25& 32.95& 25.44& 31.71& 33.74& 29.59&30.99\\ 
& Instant NGP & 29.80  & 38.24   &\cellcolor{yellow}31.75  &\cellcolor{orange}36.48  & 26.24  & 35.78  & 36.88  &\cellcolor{orange}33.03  & 33.53\\
& DVGO & \cellcolor{orange}30.23 & \cellcolor{orange}38.48 & 31.73 & 35.85 & \cellcolor{orange}26.50 & \cellcolor{orange}36.11 & \cellcolor{orange}37.26 & 32.37 & 3\cellcolor{yellow}3.57\\
& DINER & \cellcolor{yellow}30.21 & 37.90& 31.25& 35.39& 26.37& 35.13& 36.98& 32.19&33.18\\
\midrule
\multirow{7}{*}{SSIM$\uparrow$}  & RHINO (DVGO) & \cellcolor{orange}0.963  & \cellcolor{red}0.987  & \cellcolor{red}0.917  & \cellcolor{red}0.990  & \cellcolor{red}0.942  & \cellcolor{red}0.986  & \cellcolor{red}0.989  & \cellcolor{red}0.983  & \cellcolor{red}0.970 \\
& RHINO (NGP) &0.953  & 0.974  & 0.911  & \cellcolor{yellow}0.988  & 0.938  & \cellcolor{yellow}0.982  & \cellcolor{orange}0.985  & \cellcolor{red}0.983  & 0.964 \\
& NeRF &0.957  & \cellcolor{yellow}0.978  & 0.875  & 0.978  & 0.930  & 0.964  & \cellcolor{yellow}0.975  & 0.963  & 0.953 \\
& Plenoxels & 0.958  & 0.977  & 0.898  & 0.977  & 0.932  & 0.967  & 0.973  & 0.969  & 0.956\\ 
& Instant NGP & 0.956  & 0.974  & \cellcolor{yellow}0.912  & \cellcolor{yellow}0.988  & 0.938  & 0.981  & \cellcolor{orange}0.985  & \cellcolor{orange}0.982  & 0.964\\
& DVGO &\cellcolor{yellow}0.962  &\cellcolor{red} 0.987  & \cellcolor{orange}0.915  & \cellcolor{orange}0.989  & \cellcolor{yellow}0.940  & \cellcolor{red}0.986  & \cellcolor{red}0.989  & \cellcolor{orange}0.982  & \cellcolor{orange}0.969\\
& DINER & \cellcolor{red}0.964  & \cellcolor{orange}0.984  & 0.909  & \cellcolor{yellow}0.988  & \cellcolor{orange}0.941  & \cellcolor{orange}0.984  & \cellcolor{red}0.989  & \cellcolor{yellow}0.981  & \cellcolor{yellow}0.967\\
\midrule
\multirow{7}{*}{LPIPS$\downarrow$}  & RHINO (DVGO) & \cellcolor{orange}10.039  & \cellcolor{red}0.016  & \cellcolor{red}0.079  & \cellcolor{red}0.009  & \cellcolor{red}0.054  & \cellcolor{red}0.011  & \cellcolor{red}0.013  & \cellcolor{red}0.020  & \cellcolor{red}0.030 \\
& RHINO (NGP) &0.064  & 0.025  & 0.084  & \cellcolor{yellow}0.012  & 0.071  & \cellcolor{yellow}0.013  & \cellcolor{orange}0.016  & \cellcolor{orange}0.022  & 0.038 \\
& NeRF &0.044  & 0.030  & 0.142  & 0.028  & 0.075  & 0.038  &\cellcolor{yellow} 0.033  & 0.043  & 0.054 \\
& Plenoxels & 0.044  & 0.033  & 0.112  & 0.029  & \cellcolor{yellow}0.069  & 0.037  & 0.036  & 0.038  & 0.050\\ 
& Instant NGP & 0.060  & \cellcolor{yellow}0.024  & \cellcolor{yellow}0.081  & 0.014  & 0.072  & 0.014  & \cellcolor{orange}0.016  & \cellcolor{yellow}0.024  & 0.038\\
& DVGO & \cellcolor{yellow}0.040  & \cellcolor{orange}0.017  & \cellcolor{orange}0.080  & \cellcolor{red}0.009  & \cellcolor{orange}0.057  & \cellcolor{orange}0.012  & \cellcolor{red}0.013  & \cellcolor{orange}0.022  & \cellcolor{orange}0.031\\
& DINER & \cellcolor{red}0.037  & 0.025  & 0.086  & \cellcolor{orange}0.011  & \cellcolor{red}0.054  & 0.014  & \cellcolor{red}0.013  & 0.026  & \cellcolor{yellow}0.033\\\midrule
\multirow{7}{*}{Time$\downarrow$}  & RHINO (DVGO) & 407.32 & 366.77 & 483.44 & \cellcolor{orange}234.34 & \cellcolor{orange}287.59 & \cellcolor{yellow}305.44 & \cellcolor{orange}284.31 & \cellcolor{yellow}300.68 & \cellcolor{yellow}333.74 \\
& RHINO (NGP) &376.37  & 387.26  & 486.21  & 355.80  & 358.14  & 381.07  & 368.52  & 374.29  & 385.96 \\
& NeRF &10841.14  & 10841.13  & 11144.72  & 10806.78  & 10822.55  & 10847.77  & 10883.87  & 10905.11  & 10886.63 \\
& Plenoxels & \cellcolor{orange}297.88  & \cellcolor{orange}325.47  & \cellcolor{orange}360.98  & \cellcolor{yellow}277.08  & \cellcolor{yellow}289.42  & \cellcolor{orange}302.71  & \cellcolor{yellow}305.58  & \cellcolor{orange}280.80  & \cellcolor{orange}304.99\\ 
& Instant NGP & \cellcolor{yellow}335.16  & \cellcolor{yellow}349.03  & \cellcolor{yellow}400.50  & 326.25  & 352.41  & 335.63  & 324.44  & 318.67  & 342.76\\
& DVGO & \cellcolor{red}223.75  & \cellcolor{red}269.85  & \cellcolor{red}239.80  & \cellcolor{red}155.85  & \cellcolor{red}171.59  & \cellcolor{red}181.96  & \cellcolor{red}177.03  & \cellcolor{red}179.26  & \cellcolor{red}199.89 \\
& DINER & 448.54  & 467.99  & 492.14  & 272.30  & 337.73  & 467.59  & 432.39  & 422.68  & 417.67\\
\bottomrule
\end{tabularx}
\caption{Quantitative comparison on static novel view synthesis. We color code each cell as \colorbox{red}{best}, \colorbox{orange}{second best}, and \colorbox{yellow}{third best}. }
\label{tab:res_static_nerf}
\end{table*}

\subsection{Neural Radiance Fields.}

NeRF~\citep{mildenhall2020nerf} is a highly popular method for novel view synthesis. This technique renders images from novel viewpoints using multiple input images with known in/extrinsic matrixes. NeRF achieves this by mapping 3D spatial coordinates $\vec{x}$ and viewing directions $\vec{d}$ to corresponding density $\sigma$ and color emission $c$. To render the color of a pixel $\hat{C}\left( r \right) $, we firstly calculate the ray function of the $r$ using the in/extrinsic matrixes. Next, we sample $N$ points along the ray within a predefined depth range. Then, the coordinates and the directions of these $N$ samples are used to query their corresponding density and color (as performed by the MLP in NeRF). Finally, the color $\hat{C}(r)$ is obtained by applying the volume rendering techniques~\citep{max1995optical} to the queried attributes, 
\begin{equation}
\begin{aligned}
    \hat{C}(r)&=\sum_{i=1}^N T_i\left(1-\exp \left(-\sigma_i \delta_i\right)\right) c_i \\
    T_i&=\exp \left(-\sum_{j=1}^{i-1} \sigma_j \delta_j\right),
\end{aligned}
\end{equation}
where $T_i$ represents the proportion of light transmitted through ray $r$ to sample $i$ relative to the contribution of preceding samples. The term $1-\exp \left( -\sigma _i\delta _i \right)$ denotes the amount of light contributed by sample $i$, where $\sigma _i$ and $c_i$ represent the opacity and color of sample $i$, respectively. The term $\delta_i$ represents the distance between the neighboring samples. 


\begin{figure*}[!t]
\centering
\includegraphics[width=\textwidth]{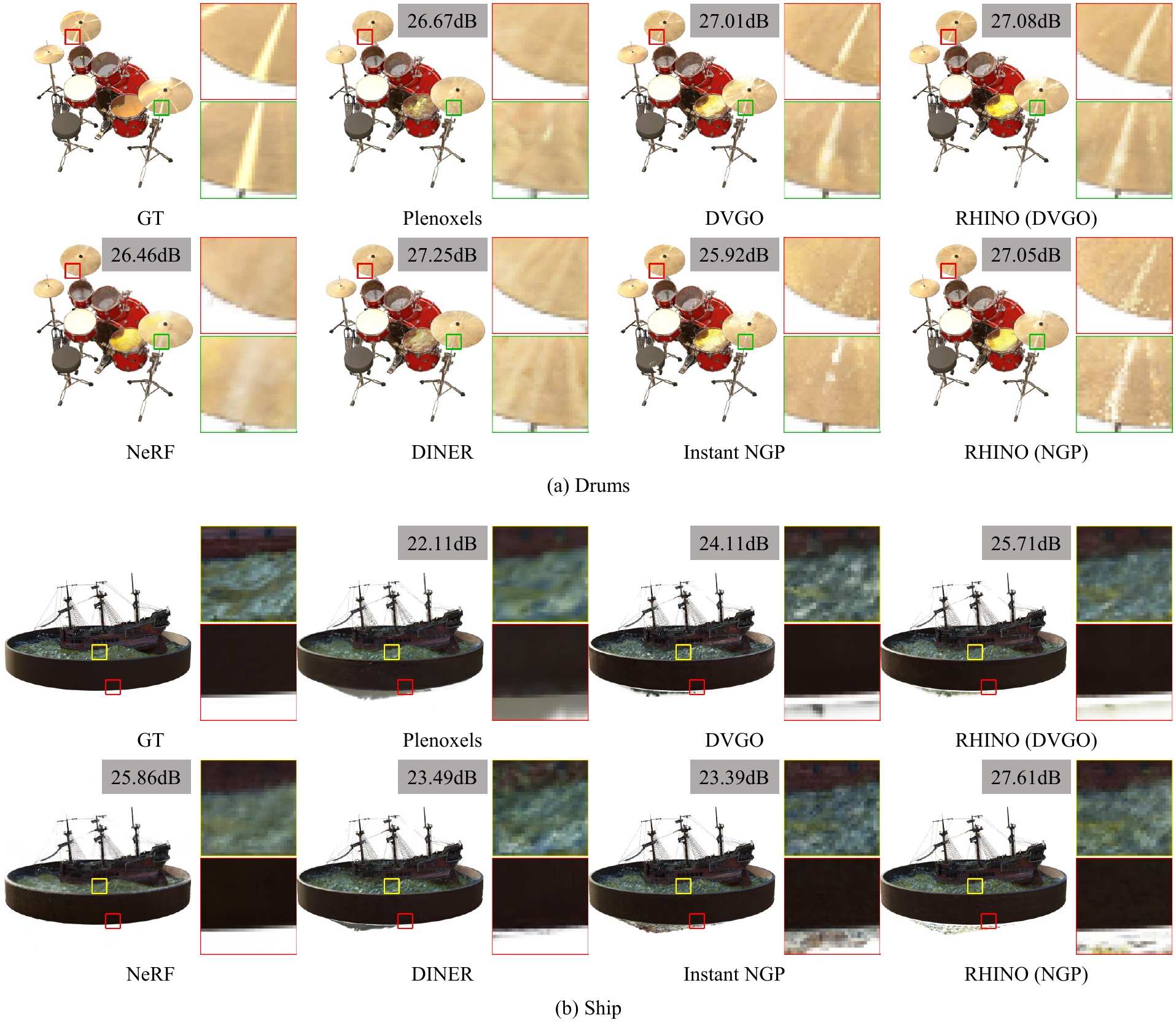} 
\caption{Qualitative comparisons of RHINO with the SOTAs on the task of the static novel view synthesis. In ’Drums’,  RHINO interpolate more reasonable textures while other methods yield smoother reconstructions. In ’Ship’, RHINO suppresses the artifacts appeared in the DVGO and the Instant NGP, producing
more smooth results in the red box.}
\label{fig:res_static_nerf}
\end{figure*}

Because the rays emitted from the input images could not cover the entire space, it is unavoidable to interpolate unsampled points when the images of novel views are rendered. As a result, it is important to design an INR with good regularization. In our experiment, five baselines are compared, namely, the NeRF, the Plenoxels~\citep{fridovich2022plenoxels}, the DINER~\citep{zhu2023disorder}, the DVGO~\citep{sun2022direct} and the Instant NGP~\citep{muller2022instant}. Among these methods, although the DINER, the DVGO and the Instant NGP all adopts the hash-based strategy, the latter two are selected for verifying the improvement of the RHINO since the DINER~\citep{zhu2023disorder} outputs the spherical harmonics coefficients~\citep{yu2021plenoctrees,fridovich2022plenoxels} instead of the attributes $(\sigma,c)$ directly.


\begin{table*}[!t]
\centering
\begin{tabularx}{0.9\textwidth}{@{}lCCCCcC@{}}
\toprule
& Scene  & Cook-Spinach & Flame-Steak &  Sear-Steak &Cut-Roasted-Beef  & Avg\\
\midrule
\multirow{4}{*}{PSNR$\uparrow$}  & Kplanes & 31.16 & 31.85 &  32.10 & \cellcolor{red}32.99 & 32.03\\
& MixVoxels &\cellcolor{orange}32.23 & 31.96 &  31.60 & \cellcolor{orange}32.94 & 32.18\\
& DINER & 31.93 & \cellcolor{orange}32.76 & \cellcolor{orange}33.07 & 32.01 & \cellcolor{orange}32.44\\
& RHINO & \cellcolor{red}32.34 & \cellcolor{red}33.51 &  \cellcolor{red}33.60 & 32.53 & \cellcolor{red}33.00\\ 
\midrule
\multirow{4}{*}{SSIM$\uparrow$}  & Kplanes & 0.932 &  \cellcolor{red}0.953  &  \cellcolor{orange}0.952 & 0.940 & 0.944\\
& MixVoxels & \cellcolor{orange}0.940 & 0.947 & 0.949  & \cellcolor{red}0.943 & \cellcolor{orange}0.945\\
& DINER & 0.936 & 0.944 & 0.948 & 0.935 & 0.941\\
& RHINO &  \cellcolor{red}0.941 & \cellcolor{orange}0.950 &  \cellcolor{red}0.953 & \cellcolor{orange}0.942 & \cellcolor{red}0.947\\
\midrule
\multirow{4}{*}{LPIPS$\downarrow$} & Kplanes & 0.192 & 0.175 &  0.178 & 0.190 & 0.184 \\
& MixVoxels & \cellcolor{red}0.140 & \cellcolor{red}0.135 &  \cellcolor{red}0.134 & \cellcolor{red}0.141 & \cellcolor{red}0.138\\
& DINER & 0.173 & 0.174 & 0.165 & 0.181 & 0.173\\
& RHINO & \cellcolor{orange}0.164 & \cellcolor{orange}0.161 & \cellcolor{orange}0.155 & \cellcolor{orange}0.170 & \cellcolor{orange}0.163\\
\midrule
\multirow{4}{*}{Time$\downarrow$} & Kplanes & 52.07 & 48.7 &  47.37 & 48.78 & 49.23 \\
& MixVoxels & 47.12 & \cellcolor{orange}44.50 &  \cellcolor{orange}45.75 & 48.42 & 46.45\\
& DINER & \cellcolor{red}40.76 & \cellcolor{red}41.52 &\cellcolor{red} 38.25 & \cellcolor{red}43.79 & \cellcolor{red}41.08\\
& RHINO & \cellcolor{orange}46.78 & 45.43 & 46.36 & \cellcolor{orange}45.35 & \cellcolor{orange}45.98\\
\bottomrule
\end{tabularx}
\caption{Quantitative comparison on dynamic view synthesis.  We color code each cell as \colorbox[RGB]{255,145,129}{best}, \colorbox[RGB]{255,188,129}{second best}. }
\label{tab:res_dyn_nerf}
\end{table*}

Tab.~\ref{tab:res_static_nerf} and Fig.~\ref{fig:res_static_nerf} offer both quantitative and qualitative comparisons of RHINO against various methods using the down-scaled Blender dataset~\citep{mildenhall2020nerf}, which has a resolution of $400\times 400$. In Tab.~\ref{tab:res_static_nerf}, the original DVGO and Instant NGP initially deliver competitive results among the five methods under consideration. However, RHINO, our proposed approach, further enhances their performance and ultimately achieves the best results. 

Fig.~\ref{fig:res_static_nerf} demonstrates these advantages well. In the example of 'Drums', the image enclosed in the red and yellow boxes demonstrate visually appealing smooth textures. Due to the broken gradients flow as mentioned above, previous hash-based methods all produce undesired artifacts in these areas which are not sampled in the training set.  RHINO rebuilds the gradients connection between the input coordinate and the output attributes, thus more reasonable textures are interpolated here. These advantages are similarly evident in the 'Ship' example, where both the waves (yellow box) and the hull bottom (red box) are sparsely sampled. RHINO effectively suppresses the artifacts observed in DVGO and Instant NGP, resulting in smoother outcomes in these under-sampled regions.


\begin{figure}[!]
\centering
\includegraphics[width=\linewidth]{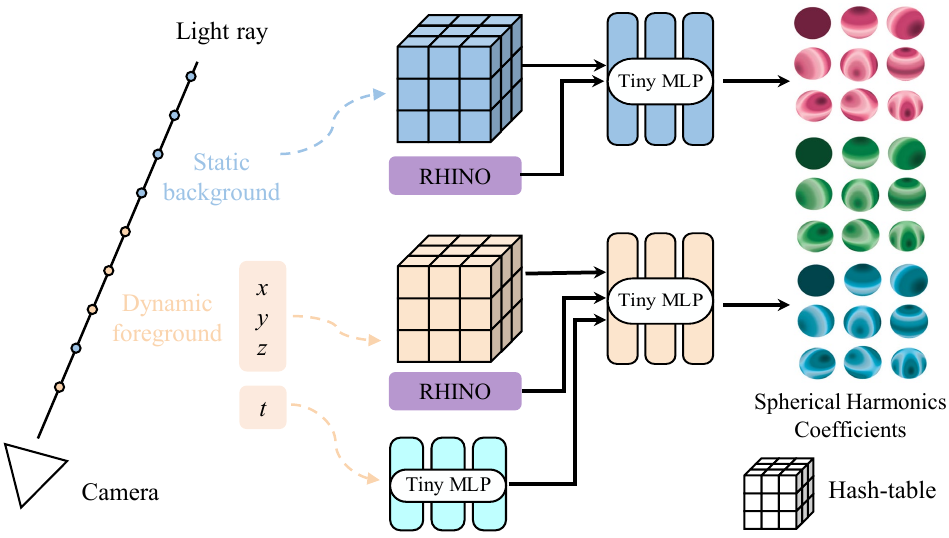} 
\caption{The pipeline of the dynamic NeRF using the RHINO.} 
\label{fig:pipeline_dynerf}
\end{figure}

\subsection{Dynamic Novel View Synthesis}

Building upon the success of NeRF for static scenes, the application of NeRF and its variants to dynamic novel view synthesis has garnered increasing attention in recent times. However, owing to the substantial volume of input data derived from multiple videos captured from different views, function-expansion-based INRs often necessitate large networks to capture intricate spatial and temporal details. This results in prohibitive time costs for training dynamic NeRF models, such as the 1300 GPU hours reported in \cite{li2022neural}. As a solution, hash-based INRs have gained widespread adoption in dynamic novel view synthesis tasks, as demonstrated by approaches like Kplanes~\citep{fridovich2023k}, Hexplanes~\citep{cao2023hexplane}, and MixVoxels~\citep{wang2022mixed}. In our implementation, we leverage RHINO with the DINER backbone~\citep{zhu2023disorder} for dynamic novel view synthesis, as depicted in Fig.~\ref{fig:pipeline_dynerf}. The 3D environment is initially partitioned into static background and dynamic foreground regions, following the methodology outlined in MixVoxels~\cite{wang2022mixed}. Subsequently, the spherical harmonic coefficients of these background and foreground points are modeled using two independent hash-based INRs. It is worth noting that instead of employing a hash-table, the time parameter $t$ is fed into an additional MLP. Finally, the view-dependent color of the 3D point is computed by querying the spherical harmonic functions based on the desired viewing direction~\citep{yu2021plenoctrees}. The size of the hash-table is set as $256^3$ with width $28$ in our implementation.

\begin{figure*}[!]
\centering
\includegraphics[width=0.95\textwidth]{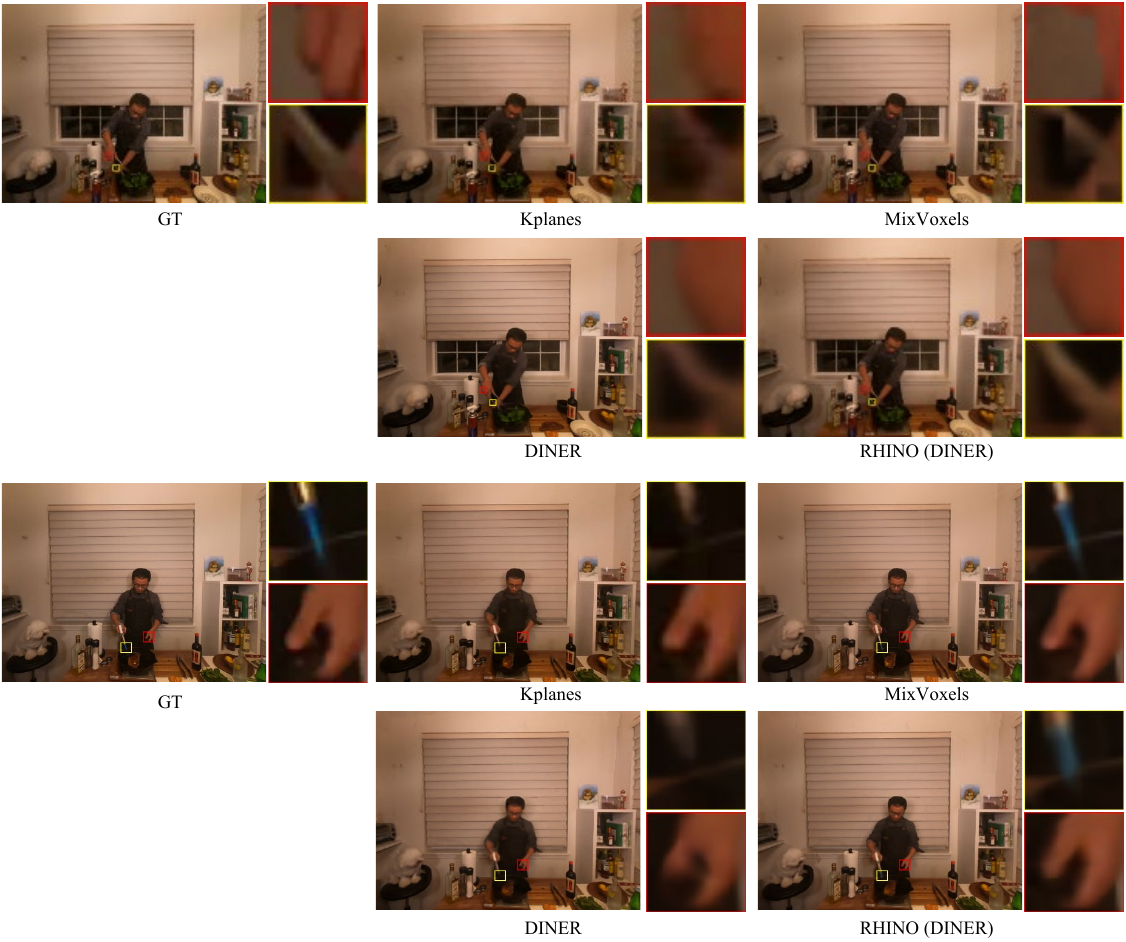} 
\caption{Qualitative comparison on dynamic novel view synthesis.} 
\label{fig:res_dyn_nerf}
\end{figure*}

We conducte our performance evaluation of RHINO using the dataset introduced by \cite{li2022neural}. This dataset comprises videos captured at a high resolution of $2028\times 2704$ pixels, with a frame rate of $30$ FPS, and encompasses $21$ distinct views. In our experiments, $10$ seconds (corresponding to $300$ frames of all images) from the $20$ views are used for training, while the frames from the remaining view were reserved for testing. To maintain consistency, we scaled the image resolution to $507\times 676$ pixels for all methods. For benchmarking purposes, we selected two SOTA methods, specifically Kplanes~\citep{fridovich2023k} and MixVoxels~\citep{wang2022mixed}, for comparison. We adhered to the default configurations outlined in their respective original implementations, following their recommended settings.


Tab.~\ref{tab:res_dyn_nerf} offers a quantitative comparison of the results. It's worth noting that RHINO outperforms the DINER backbone in all metrics, except for the additional time costs. When compared with SOTA methods, RHINO achieves the highest PSNR and SSIM scores and ranks second in terms of LPIPS values. Furthermore, RHINO demonstrates competitive training times, particularly when compared to Kplanes~\citep{fridovich2023k}, which is a popular approach for accelerating the processing of high-dimensional signals~\citep{shao2023tensor4d, singer2023text, shue20233d}.


Fig.~\ref{fig:res_dyn_nerf} offers a visual comparison of the results. Notably, in DINER, artifacts become apparent due to the broken gradient flow. For instance, observe the bent or missing handle of the spatula in Fig.~\ref{fig:res_dyn_nerf}, particularly in the zoomed-in yellow boxes found in both the top and bottom subfigures. RHINO effectively rebuilds the gradient flow, resulting in more coherent reconstructions in these problematic regions. It's noteworthy that some of the artifacts observed in DINER also persist or are exacerbated in Kplanes. This phenomenon may be attributed to the fact that in Kplanes, the input coordinates are also hash-indexed when optimizing various planes. While MixVoxels are capable of yielding clearer results, as evident in the example of the blue spatula handle in the bottom subfigure, they are occasionally prone to block artifacts. Conversely, RHINO offers more consistent reconstructions, albeit at times with the trade-off of textures appearing overly smoothed.

\section{Conclusion}
In this work, we have proposed the RHINO which is a universal method for regularizing current hash-based INRs and thus enhancing the performance of interpolations. We have pointed out that the noisy interpolations suffered in SOTA hash-based INRs are caused by the broken gradients flow between the coordinates input and the indexed hash-keys for feeding into the network. The proposed RHINO rebuilds the gradients flow by introducing a connection from the coordinate to the network directly, for which the gradients flow could be propagated from the network output to the coordinate unobstructively. For this reason, the noisy interpolation artifacts of different hash-based INRs could be significantly alleviated. Extensive experiments have verified the high accuracy and regularization performance of the proposed RHINO for various fitting and inverse problem optimization tasks.

\section*{Acknowledgements}
We thank the Prof. David B. Lindell for the help of evaluating the results of the SDF.

\section*{Data Availability Statements}
The datasets used for the image representation, 3D shape representation, as well as the 5D static/6D dynamic NeRF come from the public datasets~\citep{asuni2014testimages, standord-3D-scanning,mildenhall2020nerf,li2022neural}. All codes and models will be publicly available to the research community to facilitate reproducible research once the paper is accepted. The datasets generated during and/or analysed during the current study are available from the corresponding author on reasonable request.

\bibliographystyle{spbasic}
{\footnotesize \bibliography{ref.bib}}


\end{document}